\documentclass[letterpaper]{article}
\usepackage[preprint]{aaai2027}
\usepackage[hyphens]{url}
\usepackage{graphicx}
\usepackage{natbib}
\usepackage{booktabs}
\title{Agentic Building-Aware Satellite Gaussian Splatting for Auditable Urban DSM Reconstruction}
\author{
    Wentao Sun\textsuperscript{\rm 1},
    Zhengsen Xu\textsuperscript{\rm 2},
    Yiping Chen\textsuperscript{\rm 3},\\
    John S. Zelek\textsuperscript{\rm 1},
    Jonathan Li\textsuperscript{\rm 1}
}
\affiliations{
    \textsuperscript{\rm 1}University of Waterloo, Department of Systems Design Engineering, Waterloo, Canada\\
    \textsuperscript{\rm 2}University of Calgary, Department of Geomatics Engineering, Calgary, Canada\\
    \textsuperscript{\rm 3}Sun Yat-sen University, School of Geospatial Engineering and Science, Zhuhai, China\\
    wentao.sun@uwaterloo.ca, chenyp79@mail.sysu.edu.cn, zhengsen.xu@ucalgary.ca, jzelek@uwaterloo.ca, junli@uwaterloo.ca
}

\begin{document}
\maketitle

\begin{abstract}
Urban-scale 3D reconstruction from satellite imagery supports disaster response, city monitoring, and geospatial digital twins, yet neural rendering methods typically optimize average visual fidelity rather than the structures that analysts inspect first: buildings. We present an agentic building-aware satellite Gaussian Splatting workflow that uses Segment Anything-derived building masks as semantic priors and an Agentic Reconstruction Controller to select, verify, and record DSM reconstruction policies. On the DFC2019 JAX\_004 scene, building-aware weighting reduces building-region DSM MAE from 0.844 m to 0.806 m, showing that semantic priors can shift reconstruction capacity toward analyst-critical regions. A staged schedule provides a balanced operating point, improving full-scene MAE from 1.362 m to 1.349 m while retaining a building gain. Across four JAX scenes, the Agent selects validated policies for both general DSM and building-focused DSM objectives, and produces building-inventory metadata and per-scene decision records. The system combines semantic priors, policy selection, region-specific DSM metrics, and DSM-derived GIS surface products for auditable urban 3D analysis.
\end{abstract}

\begin{links}
    \link{Project Page}{https://w27sun.github.io/agentdsm/}
\end{links}

\section{Introduction}
Satellite image collections offer wide-area coverage for cities, disaster zones, and changing infrastructure, but converting those images into reliable 3D products remains difficult. Operational users often need DSMs, roof-level cues, and inspectable 3D surfaces from imagery collected under heterogeneous viewing angles, seasonal effects, shadows, and changing illumination. Modern neural rendering and Gaussian Splatting methods can represent complex appearance, yet their optimization is usually driven by image reconstruction losses rather than by the semantic structures that geospatial users inspect first. In urban analysis, a meter-level improvement on buildings can matter more than a visually plausible texture on roads, parking lots, or vegetation.

This paper asks a practical question: can foundation-model semantic masks and an Agentic Reconstruction Controller make satellite Gaussian Splatting more useful for operational DSM reconstruction? We focus on buildings because they are visible in satellite imagery, application-critical, and measurable with public truth layers. The method is designed as a deployable batch workflow: generate per-view SAM-style building masks~\cite{kirillov2023segment}, inject them as photometric weights during Earth-observation Gaussian Splatting (EOGS), render DSMs, evaluate semantic-region errors, and export analyst-facing visual products.

The main finding is building-centric. On JAX\_004, semantic weighting reduces building MAE from 0.844 m to 0.806 m. A staged schedule gives a balanced operating point that improves full-scene MAE while retaining a building gain. Additional JAX scenes are used to evaluate an Agent-guided readiness policy: the controller observes mask quality, inventory metadata, and validation metrics, then selects building-aware enhancement or the baseline operating mode according to the requested product. The contribution is therefore a deployable decision workflow for urban DSM production, not a single global semantic weight.

The paper makes four application-oriented contributions. First, it frames foundation segmentation as a task prior for satellite Gaussian Splatting rather than as a stand-alone segmentation product. Second, it introduces a region-weighted photometric objective and staged semantic-weight schedule for building-aware EOGS optimization. Third, it adds an Agentic Reconstruction Controller that turns mask QA, validation metrics, and building-inventory metadata into recorded policy decisions. Fourth, it evaluates DSM error by application region and exports analyst-facing DSM, GIS surface previews, and decision records.

\section{Related Work and Positioning}
Urban 3D products support planning, energy analysis, disaster management, visualization, and digital-twin maintenance across many city-model use cases~\cite{biljecki2015applications,ketzler2020digitaltwins}. Classical satellite 3D reconstruction has long relied on stereo and multi-view photogrammetry, with DFC2019/US3D providing a widely used benchmark for large-scale semantic 3D reconstruction from incidental satellite imagery~\cite{bosch2019semantic,lesaux2019dfc}. Neural radiance fields introduced continuous scene representations for view synthesis~\cite{mildenhall2020nerf}, and satellite-specific variants such as Sat-NeRF and EO-NeRF model RPC cameras, shadows, and transient appearance to improve DSM recovery from multi-date imagery~\cite{mari2022satnerf,mari2023eonerf}. Efficiency-oriented follow-ups, including SAT-NGP and EOGS, reduce training time and adapt neural rendering or Gaussian Splatting to Earth observation~\cite{billouard2024satngp,aira2025eogs}. Recent satellite Gaussian Splatting work further explores generalizable sparse-view reconstruction and semantic feature fusion~\cite{huang2026skysplat,reed2026semanticgs}.

Foundation segmentation models provide a complementary direction. SAM introduced promptable segmentation at broad scale~\cite{kirillov2023segment}; subsequent remote-sensing studies demonstrate its promise for overhead imagery and motivate domain-specific adaptation~\cite{ren2024samspace,osco2023samrs,wu2023samgeo}. Building-specific SAM adaptations improve footprint extraction and boundary quality~\cite{li2025buildingsam,wei2024sampolybuild}, but they primarily evaluate 2D segmentation. Our work instead uses building masks as task priors for 3D satellite Gaussian Splatting and couples them with a scene-level operating policy.

Recent language-agent work provides a useful abstraction for this operating policy. ReAct-style systems interleave reasoning with tool actions~\cite{yao2023react}, Reflexion uses feedback from previous attempts~\cite{shinn2023reflexion}, and multi-agent frameworks such as AutoGen organize planner, executor, and verifier roles~\cite{wu2023autogen}. We adapt this Agent pattern to geospatial reconstruction: the Agent does not replace the renderer, but observes mask QA and DSM validation outputs, selects reconstruction policies, and writes a decision record for analyst review.

Mesh extraction is also related to the downstream product goal. Surface-aligned Gaussian Splatting and 2D Gaussian Splatting show that Gaussian representations can support explicit surfaces and editable meshes~\cite{guedon2024sugar,huang2024twodgs}. Recent satellite Gaussian variants further improve season handling, shadow modeling, sparse-view robustness, and surface reconstruction~\cite{xu2026sags,luo2026shadowgs,kim2026geogs,chen2026satsurf}. In contrast to full surface-aligned retraining, our system treats the satellite DSM/altitude output as the reliable deployment geometry and visualizes it with hillshade, building-footprint overlays, and DSM-derived oblique surface previews, leaving stronger surface regularization as future work.

\begin{table}[t]
\centering
\scriptsize
\setlength{\tabcolsep}{2.5pt}
\renewcommand{\arraystretch}{0.88}
\begin{tabular}{p{0.24\columnwidth}p{0.32\columnwidth}p{0.32\columnwidth}}
\toprule
Family & Main emphasis & Difference in this work \\
\midrule
Sat-NeRF / EO-NeRF & RPC NeRF, shadows, DSM/NVS & Adds building-task prior and Agent policy \\
SAT-NGP / EOGS & Efficient satellite neural/GS reconstruction & Uses EOGS as engine, adds semantic product control \\
Recent satellite GS & Season, shadow, sparse-view, or generalization modules & Selects AOI-specific building-aware DSM products \\
Surface-oriented GS & Explicit surface and mesh extraction & Reports DSM-derived GIS surface previews and audit records \\
\bottomrule
\end{tabular}
\caption{External method positioning. The comparison is by system role rather than a direct SOTA leaderboard because the proposed contribution is a semantic-prior and Agent-control layer around EOGS.}
\label{tab:external-positioning}
\end{table}

Table~\ref{tab:external-positioning} positions the work as a semantic-prior and policy-control layer around EOGS. The application gap is turning foundation segmentation into a controlled reconstruction prior for metric satellite geometry, with AOI-specific reporting instead of one global hyperparameter.

\section{Operational Setting}
The target user is a geospatial analyst or urban digital-twin engineer who needs a registered DSM plus interpretable 3D evidence from multi-view satellite imagery. Because such users inspect GIS-ready products rather than neural representations, the workflow treats Gaussian Splatting as a reconstruction engine inside a product pipeline.

The application is also asymmetric: not every pixel is equally important. Building roofs, roof edges, and building blocks are often the first regions inspected in post-disaster mapping, urban growth monitoring, and infrastructure inventory. Vegetation can dominate height error in many scenes, but it is less stable across dates and less likely to support crisp planar geometry. A single full-scene MAE can therefore hide the effect that matters to the user. The central design choice in this paper is to make the optimization and the evaluation region-aware.

The deployment constraints are practical: no manual polygon annotation per AOI, inspectable mask diagnostics, a retained baseline mode, and artifacts viewable without neural-rendering tools. These constraints motivate soft photometric weights rather than hard geometric constraints.

\begin{figure}[t]
\centering
\includegraphics[width=\columnwidth]{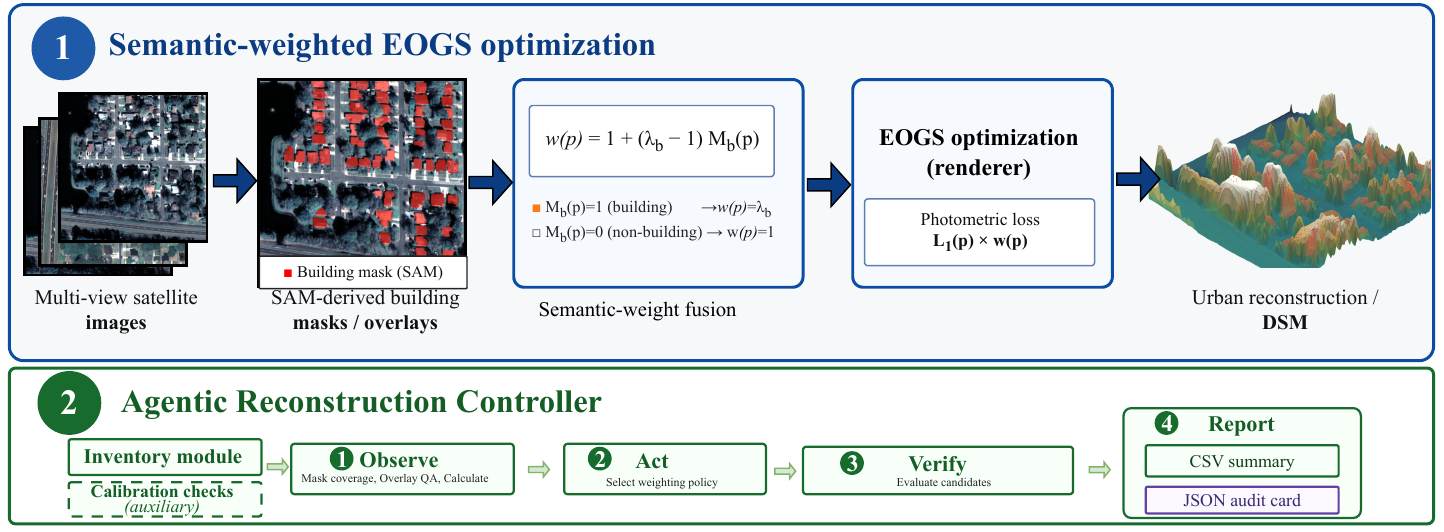}
\caption{System overview. Multi-view satellite images and SAM-derived building masks feed candidate EOGS reconstructions. The Agentic Reconstruction Controller observes mask QA and inventory metadata, acts by selecting a policy, verifies region-specific DSM metrics, and reports an auditable product bundle.}
\label{fig:pipeline-placeholder}
\end{figure}

Figure~\ref{fig:pipeline-placeholder} summarizes the intended workflow. Given multi-view satellite images and camera models, a SAM-family segmenter produces per-view building masks. EOGS optimizes a Gaussian scene representation with a weighted photometric objective. The Agentic Reconstruction Controller sits around this renderer: it observes mask coverage and inventory signals, chooses the candidate schedule to run, verifies full-scene and region-specific DSM metrics, and records a decision trace. The trained representation then supports rendered images, DSM extraction, DSM error maps, and DSM-derived GIS surface visualization.

\begin{table}[t]
\centering
\scriptsize
\setlength{\tabcolsep}{3pt}
\renewcommand{\arraystretch}{0.92}
\begin{tabular}{p{0.25\columnwidth}p{0.60\columnwidth}}
\toprule
Deployment item & System contract \\
\midrule
Target user & Geospatial analyst inspecting urban DSMs and 3D surfaces \\
Input & Multi-view satellite images, camera models, optional SAM-derived masks \\
Output & Registered DSM, building-region errors, inventory metadata, decision record, GIS surface preview \\
Agent observations & Mask coverage, inventory metadata, overlay QA, canonical-view DSM validation \\
Agent action & Select baseline, staged, or building-specialist semantic schedule \\
Agent verifier & Report selected policy, full/building/non-building metrics, and decision trace \\
\bottomrule
\end{tabular}
\caption{Deployment contract for the system workflow.}
\label{tab:deployment-contract}
\end{table}

Table~\ref{tab:deployment-contract} states the system contract: observe mask plausibility and inventory, select a candidate EOGS schedule, verify DSM metrics, and report the selected operating mode.

\section{Agentic Semantic-Prior EOGS}
Let $L_1(p)$ denote the per-pixel photometric reconstruction loss and $M_b(p)$ the SAM-derived building-mask value for pixel $p$. The baseline EOGS objective is modified by a per-pixel weight
\[
    w(p) = 1 + (\lambda_b - 1)M_b(p),
\]
so building-labeled pixels receive stronger photometric supervision. In practice, the masks are generated offline for each training image and read during optimization. This keeps the EOGS renderer, camera model, and DSM evaluation pipeline unchanged; only the loss weighting changes. If no mask is provided, or if $\lambda_b=1$, the method reduces to the original baseline.

The weighting has an application-oriented interpretation: a building mask does not assert known height, but marks pixels whose photometric residuals matter more to the product. The mask can therefore be soft, imperfect, and view-dependent, which is important under satellite shadows, occlusion, and multi-date appearance changes.

We evaluate static and staged weights. Static weights characterize the building-vs.-global trade-off; the staged setting increases the weight from 1.3 to 1.5 to 2.0 within the same 5000-iteration budget, first establishing broad geometry and then emphasizing buildings. Stronger weights such as 2.5 test building-specialist settings.

The second part of the method is a mask-quality gate. For a scene $s$, we compute a mean mask coverage score
\[
    c_s = \frac{1}{N_s}\sum_i \frac{\sum_{p \in \Omega_i} M_i(p)}{|\Omega_i|},
\]
where $N_s$ is the number of training views. Coverage is not used as a formal proof of correctness; it is an early quality signal. High coverage can suggest that the segmenter may be labeling broad hard surfaces, roads, or urban texture as buildings. Low coverage is therefore paired with validation before selecting aggressive building-specialist weights.

The Agentic Reconstruction Controller turns this gate into an executable policy with four steps. Observe computes mask coverage, overlay QA, building coverage/count/area/relief, and candidate-validation DSM metrics. Act selects baseline, staged, or building-specialist candidates under the requested product objective. Verify records full-scene, building, vegetation, and non-building MAE for the selected policy. Report writes a CSV summary and JSON audit card, so the decision can be inspected independently of the renderer. In the current offline run, the action step uses candidate-set validation metrics; we do not treat it as a learned policy that generalizes without validation.

The inventory module reports connected building components, footprint area, and DSM relief as analyst-facing metadata rather than cadastral labels. Vegetation downweighting and depth-prior losses serve as calibration checks; the controller centers building masks because they best match the urban DSM product.

\section{Experimental Setup}
Experiments use DFC2019/IARPA-style satellite scenes with ground-truth DSM and semantic class maps. We report full-scene DSM MAE and semantic-region MAE for buildings (CLS=6), vegetation (CLS=5), and non-building pixels. Predicted DSMs are registered to ground truth using the existing EOGS DSM evaluation pipeline before computing errors. The metric is simple by design: it matches the kind of height error that downstream mapping users inspect, and it can be computed for every ablation without requiring a new learned evaluator.

The primary semantic experiments use SAM-derived building masks generated offline for each training view, without manual mask correction. All reported semantic-prior runs use 5000 EOGS iterations and the same train/test split as the baseline. We evaluate a controlled JAX\_004 ablation and cross-scene staged experiments on JAX\_068, JAX\_214, and JAX\_260; canonical DSM test indices are fixed per scene before evaluation. Baseline multi-scene EOGS reproduction metrics provide context for how scene difficulty varies before semantic priors are added.

JAX\_004 is the primary building-enhancement scene because its building mask is visually plausible and its baseline building error leaves room for improvement. JAX\_068 and JAX\_214 provide high-coverage cases for testing the readiness policy, while JAX\_260 provides a low-coverage case with a strong baseline. Together, these scenes support both the positive building result and the policy-selection analysis used for deployment. The Agent report script uses the same completed metric tables, SAM-mask coverage logs, and benchmark CLS/DSM rasters to generate inventory metadata and policy decision records.

\begin{table}[t]
\centering
\scriptsize
\setlength{\tabcolsep}{2pt}
\renewcommand{\arraystretch}{0.86}
\resizebox{\columnwidth}{!}{%
\begin{tabular}{llrrrr}
\toprule
Scene & Variant & Full MAE & Bldg. MAE & Veg. MAE & Non-bldg. MAE \\
\midrule
JAX\_004 & Baseline & 1.3625 & 0.8440 & 3.2068 & 1.4238 \\
JAX\_004 & SAM 1.3 & 1.3615 & 0.8326 & 3.2307 & 1.4241 \\
JAX\_004 & SAM 2.0 & 1.3757 & \textbf{0.8061} & 3.1939 & 1.4432 \\
JAX\_004 & SAM 2.5 & 1.3515 & 0.8077 & 3.1493 & 1.4159 \\
JAX\_004 & Staged & \textbf{1.3486} & 0.8347 & 3.1710 & \textbf{1.4094} \\
JAX\_068 & Baseline & \textbf{1.0926} & \textbf{1.0478} & \textbf{1.6069} & \textbf{1.1266} \\
JAX\_068 & Staged & 1.1038 & 1.0685 & 1.7942 & 1.1305 \\
JAX\_068 & S1.1 & 1.1039 & 1.0609 & 1.7362 & 1.1365 \\
JAX\_214 & Baseline & \textbf{1.7493} & \textbf{1.3635} & 3.0016 & 2.1635 \\
JAX\_214 & Staged & 1.7541 & 1.4252 & \textbf{2.7626} & \textbf{2.1072} \\
JAX\_214 & S1.1 & 1.7589 & 1.3777 & 3.1045 & 2.1680 \\
JAX\_260 & Baseline & \textbf{1.5502} & \textbf{0.8303} & \textbf{2.2945} & \textbf{1.6452} \\
JAX\_260 & Staged & 1.5726 & 0.8665 & 2.3114 & 1.6658 \\
JAX\_260 & S2.5 & 1.7282 & 0.9527 & 2.3232 & 1.8305 \\
\bottomrule
\end{tabular}%
}
\caption{DSM MAE in meters. JAX\_004 shows the building-region gain, while the additional scenes support readiness-policy selection for semantic weighting.}
\label{tab:key-results}
\end{table}

\section{Results}
Table~\ref{tab:key-results} shows the central trade-off. On JAX\_004, a static weight of 2.0 gives the best building-region MAE, improving from 0.844 m to 0.806 m, but it prioritizes building accuracy over full-scene error. A stronger static weight of 2.5 is nearly tied on buildings at 0.808 m and lowers full-scene MAE to 1.352 m, making it a useful deployment compromise. The staged schedule is the best current global compromise, improving full MAE from 1.362 m to 1.349 m and non-building MAE from 1.424 m to 1.409 m while still improving building error relative to baseline.

\begin{figure*}[t]
\centering
\begin{minipage}{0.48\textwidth}
\centering
\includegraphics[width=\linewidth]{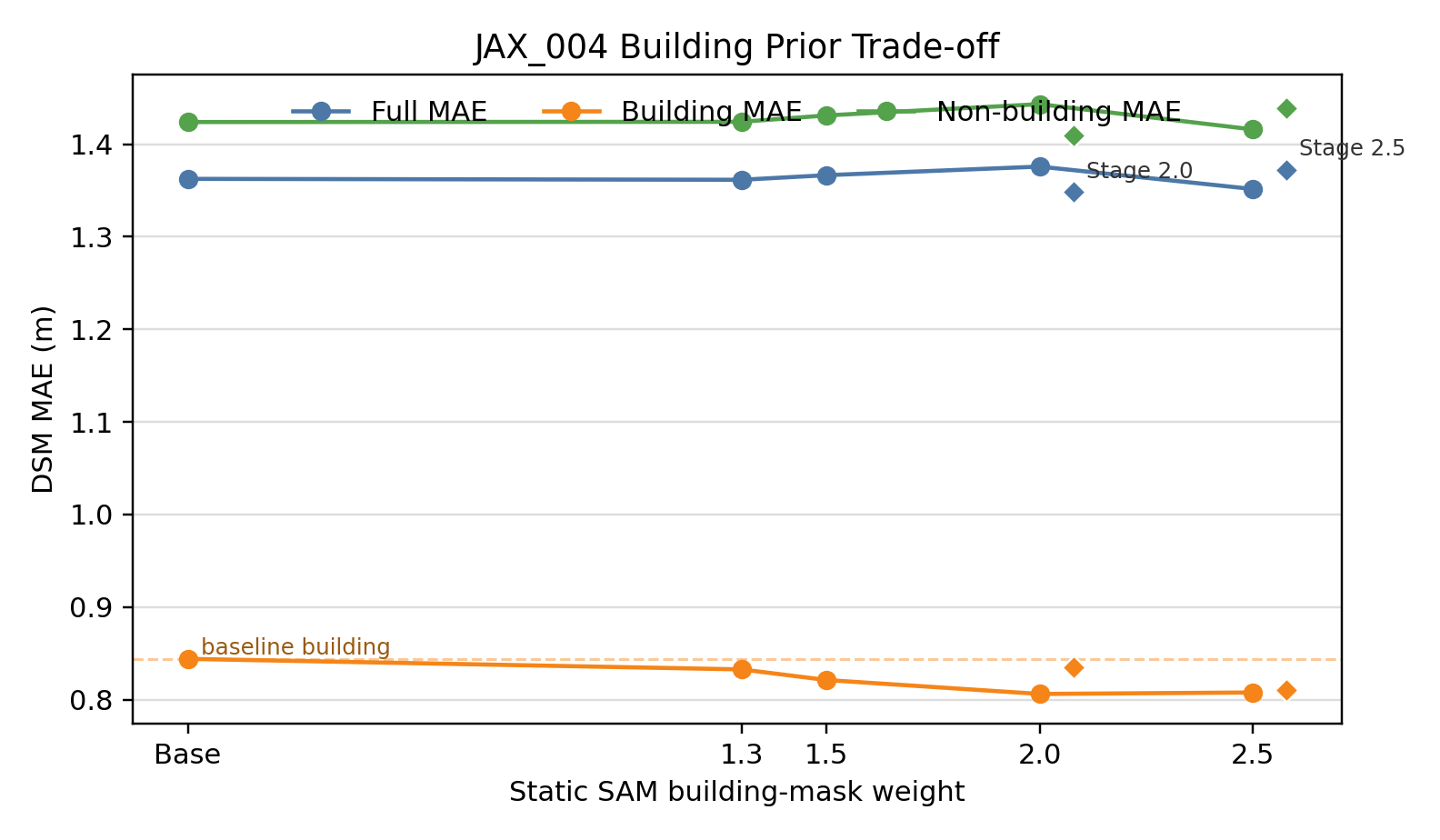}
\end{minipage}
\hfill
\begin{minipage}{0.48\textwidth}
\centering
\includegraphics[width=\linewidth]{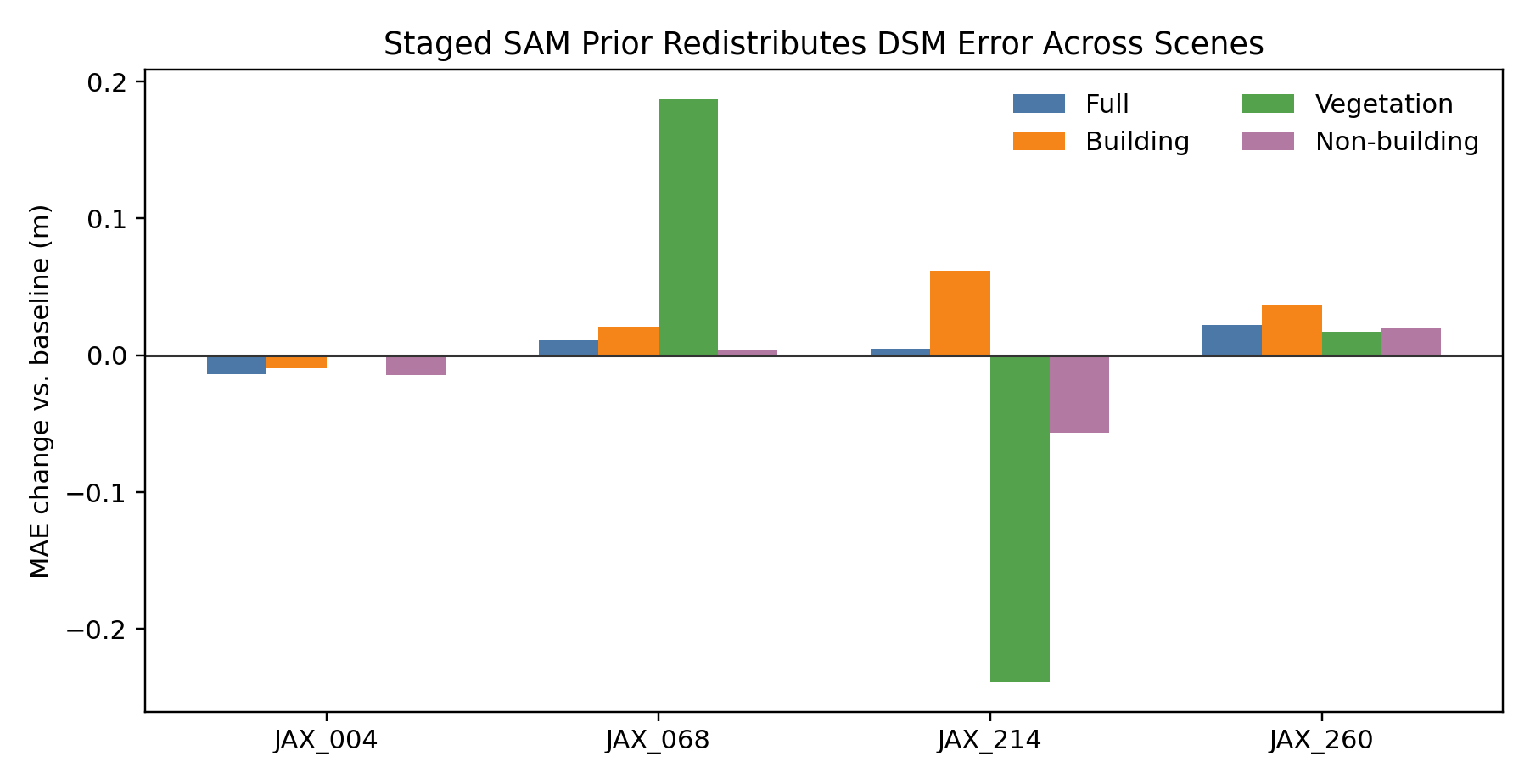}
\end{minipage}
\caption{Building/global operating points and cross-scene regional DSM deltas. Deltas below zero indicate improvement over each baseline.}
\label{fig:tradeoff-delta}
\label{fig:building-tradeoff}
\label{fig:delta}
\end{figure*}

The JAX\_004 ablation shows distinct operating points. Weight 1.3 is conservative, improving buildings with little global change. Weight 2.0 is the building-specialist setting, giving the best building MAE with a stronger building-priority trade-off. Weight 2.5 retains near-best building accuracy while improving full-scene MAE relative to weight 2.0. The staged setting favors full-scene and non-building MAE while preserving a smaller building gain. This supports policy selection because deployment settings may prefer different trade-offs depending on the user task.

The cross-scene results motivate policy selection rather than a fixed semantic weight. JAX\_004 selects the staged building-aware mode, while JAX\_068, JAX\_214, and JAX\_260 select baseline or conservative modes under the readiness policy. This behavior is useful in an applied reconstruction system: semantic priors are activated when they improve the requested product, and the baseline remains available as a high-quality operating point. Figure~\ref{fig:tradeoff-delta} visualizes the regional error deltas used by this policy.

\begin{figure*}[t]
\centering
\begin{minipage}{0.43\textwidth}
\centering
\includegraphics[width=\linewidth]{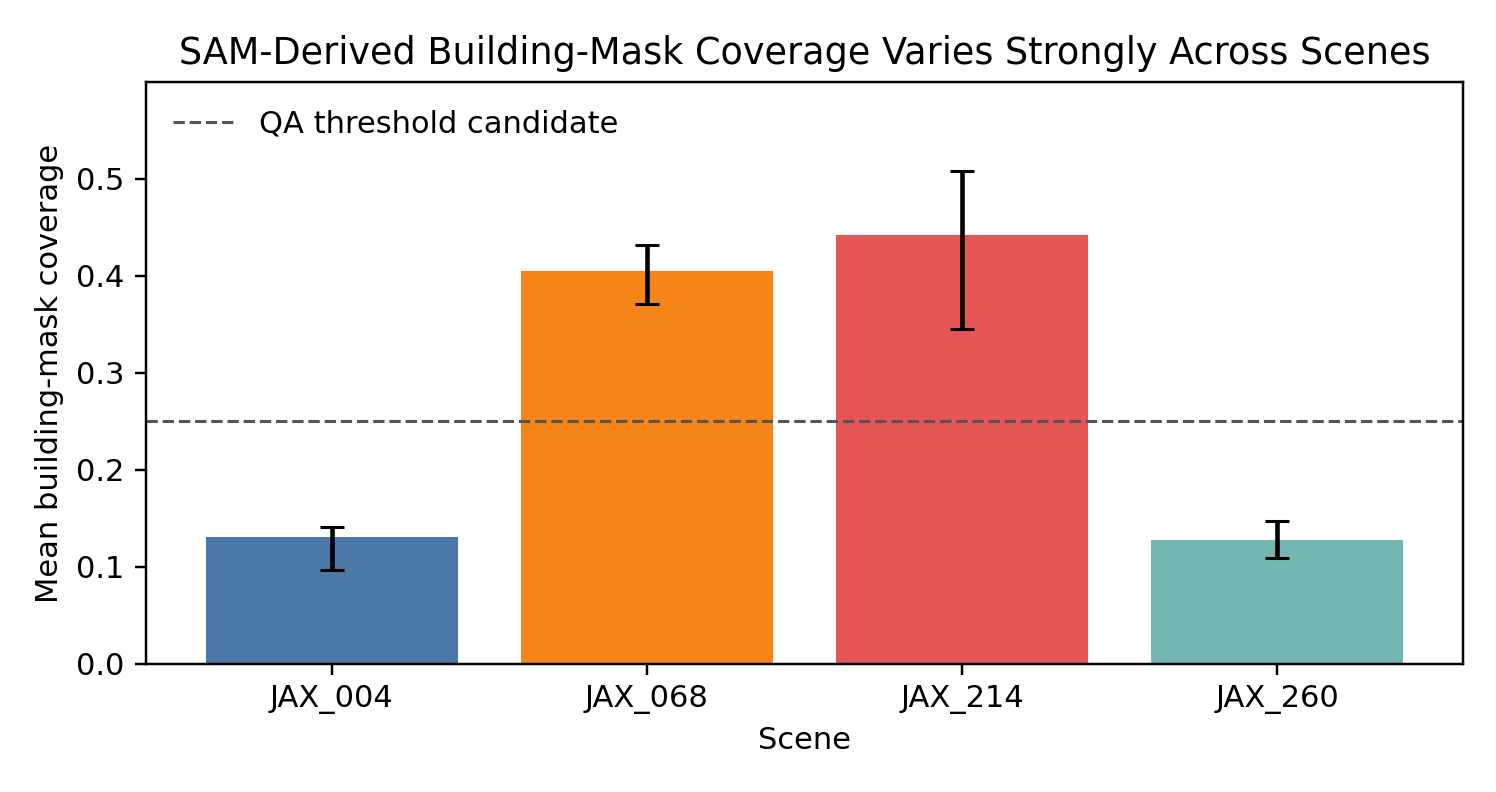}
\end{minipage}
\hfill
\begin{minipage}{0.55\textwidth}
\centering
\includegraphics[width=\linewidth]{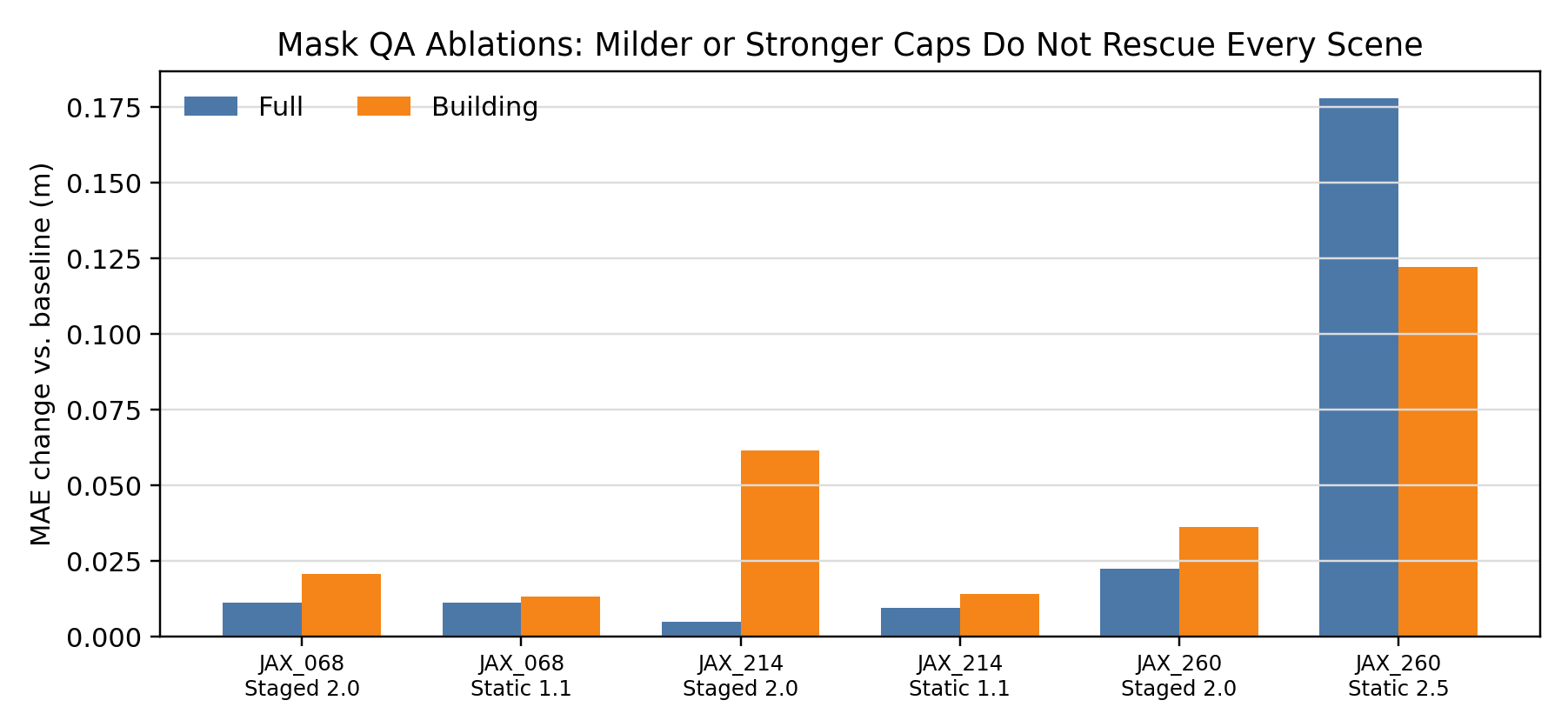}
\end{minipage}
\caption{Mask-QA diagnostics. Coverage provides an interpretable prior-quality signal, and validation ablations complete the operating-policy decision.}
\label{fig:mask-qa}
\label{fig:coverage}
\label{fig:qa-caps}
\end{figure*}

Figure~\ref{fig:mask-qa} shows how mask coverage and validation complement each other. JAX\_068 and JAX\_214 have mean building-mask coverage near 0.405 and 0.442, much higher than JAX\_004 and JAX\_260, so the readiness policy treats them as candidates for conservative operation. JAX\_260 illustrates the value of the validation step: even with low coverage, the baseline can remain the preferred product setting. The resulting gate combines coverage, overlay inspection, and a cheap validation run.

The auxiliary-prior experiments further support the design choice. The vegetation downweight experiment was motivated by the fact that trees and seasonal vegetation can be unstable across views; it serves as a calibration check for whether semantic reweighting should target all unstable regions or the application-critical region. The current results favor building masks because they align directly with the target geometry, the sensor evidence, and the metric evaluation. These comparisons sharpen the contribution: the useful prior is not ``any foundation model output,'' but a prior that matches the task region, the sensor geometry, and the requested product.

\subsection{Agent-Generated Reconstruction Report}
To make the readiness policy measurable, we run an offline Agentic Reconstruction Controller over the four-scene result set. The Agent receives the candidate-validation metrics in Table~\ref{tab:key-results}, SAM-mask coverage, and AOI inventory metadata computed from the benchmark building layer. It then selects two products per AOI: a general DSM policy that minimizes full-scene MAE, and a building-focused policy that minimizes building-region MAE. A policy-regret check shows the value of AOI-specific selection: for the general DSM objective, the Agent selects a mean MAE of 1.4352, compared with 1.4386 for fixed baseline and 1.4448 for fixed staged operation; for the building-focused objective, it selects 1.0119, compared with 1.0214 and 1.0487. The run also writes per-scene decision records to disk, recording the observations, selected policies, and metrics used by the decision.

\begin{table*}[t]
\centering
\scriptsize
\setlength{\tabcolsep}{3.5pt}
\renewcommand{\arraystretch}{0.88}
\begin{tabular}{lrrrrrlr}
\toprule
Scene & Bldg. area & Components & SAM cov. & General policy & Full MAE & Building policy & Bldg. MAE \\
\midrule
JAX\_004 & 6,910 m$^2$ / 10.5\% & 63 & 0.131 & Staged & \textbf{1.3486} & Static 2.0 & \textbf{0.8061} \\
JAX\_068 & 28,222 m$^2$ / 43.1\% & 9 & 0.405 & Baseline & \textbf{1.0926} & Baseline & \textbf{1.0478} \\
JAX\_214 & 32,959 m$^2$ / 50.3\% & 3 & 0.442 & Baseline & \textbf{1.7493} & Baseline & \textbf{1.3635} \\
JAX\_260 & 5,626 m$^2$ / 8.6\% & 22 & 0.127 & Baseline & \textbf{1.5502} & Baseline & \textbf{0.8303} \\
\bottomrule
\end{tabular}
\caption{Agent-generated AOI reconstruction report. Building area and component count are benchmark-grid inventory metadata (CLS=6, 0.5 m grid, connected components $\ge$25 m$^2$). The Agent selects the validated policy for both general DSM and building-focused DSM objectives on all four scenes.}
\label{tab:agent-report}
\end{table*}

Table~\ref{tab:agent-report} reframes the multi-scene experiment as product selection. For JAX\_004, the controller selects Staged for the best general DSM and Static 2.0 for the best building DSM. For JAX\_068, JAX\_214, and JAX\_260, it keeps the validated baseline product. In all cases, the output is a recorded policy decision rather than a manually chosen hyperparameter.

The inventory fields add operational metadata--footprint area, roof-block count, mask coverage, selected policy, and region-specific error--that make the product easier to triage, compare, and archive.

\begin{figure*}[t]
\centering
\includegraphics[width=0.72\textwidth]{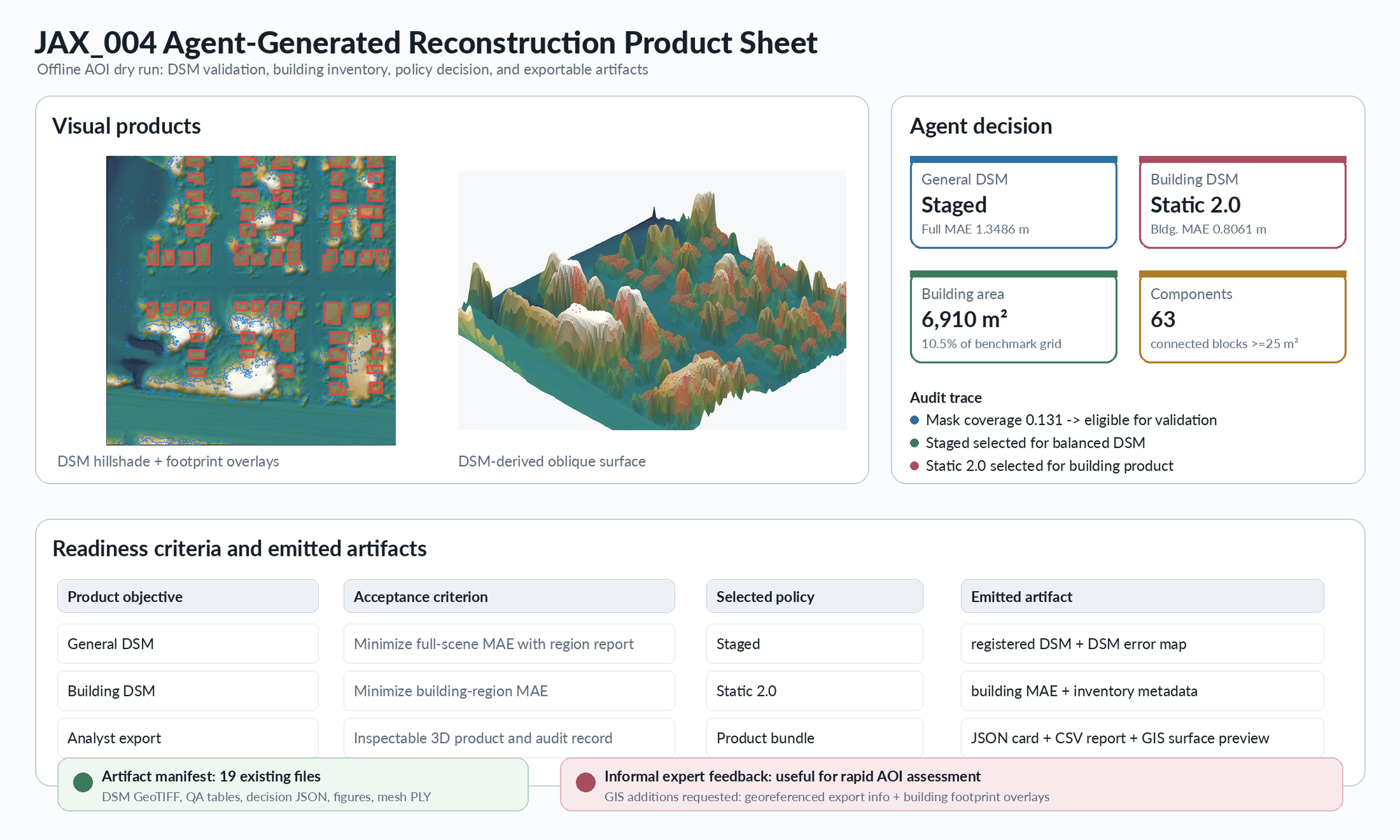}
\caption{Agent-generated JAX\_004 product sheet combining DSM hillshade overlays, a DSM-derived oblique surface preview, AOI inventory, selected policies, readiness criteria, and emitted artifacts.}
\label{fig:product-outputs}
\end{figure*}

\section{Deployment Lessons}
Figure~\ref{fig:product-outputs} summarizes the intended product view. For JAX\_004, the offline AOI run emits a registered DSM, DSM error view, building inventory, selected general and building policies, JSON/CSV decision records, a hillshade overlay, and a DSM-derived oblique surface preview. The key design choice is that the selected policy is reported with the evidence that justified it. Analysts can therefore inspect not only the final surface but also the mask coverage, inventory context, and regional DSM errors that led the Agent to select staged, static, or baseline operation.

The experiments suggest three deployment lessons. First, foundation-model masks can steer reconstruction without retraining the segmenter or the renderer, but the gain is scene dependent. Second, visually plausible masks are not sufficient for metric geometry; strong semantic weighting should be selected only after DSM validation confirms that the requested region improves. Third, auxiliary priors must match the product objective. In these experiments, building masks are useful because the region is visible, measurable, and operationally important; vegetation downweighting and monocular depth priors require stronger confidence calibration before deployment.

Operationally, the Agent should run a QA gate before full training: compute mask coverage, inspect mask overlays, summarize building inventory, and run cheap canonical-view validation before schedule selection. High-coverage scenes are routed to baseline or conservative settings unless validation supports stronger building weights; low-coverage scenes can advance to stronger settings such as 2.5 when the requested product prioritizes roof accuracy. A deployment dashboard should report the selected weight, selection rationale, inventory metadata, full-scene MAE, building MAE, and non-building MAE so the trade-off is visible instead of hidden behind a single aggregate score.

The adoption path is incremental. A first product release can provide DSMs, error maps, mask overlays, DSM-derived surface previews, and an artifact manifest. Later runs can add cached inventory, validation metrics, and policy histories across AOIs. This keeps the learning-based renderer replaceable: future satellite Gaussian Splatting engines, stronger segmentation models, or surface-aligned mesh extractors can be inserted while preserving the same validation record and analyst-facing decision contract.

\section{Scope and Future Work}
This study focuses on four JAX scenes, building-focused masks, DSM-region metrics, and Agent-generated decision records. The scope is sufficient to test the proposed operating policy, but not to claim a universal semantic weighting rule. JAX\_004 establishes the building-aware enhancement case, while the other scenes show why policy selection matters under different mask coverage, inventory, and baseline-strength conditions.

Future work should add calibrated mask-quality prediction from coverage, connected-component statistics, cross-view agreement, rendered-geometry alignment, and validation loss. The weighting schedule should also become adaptive: rather than choosing from a fixed set of 1.3, 2.0, and 2.5 settings, the controller should increase building emphasis only while a full-scene accuracy bound is satisfied. This would turn the current candidate-selection procedure into a more scalable AOI policy.

A second direction is stronger surface export. The current product uses the DSM/altitude output as dependable geometry and avoids treating normalized Gaussian RGB as a photorealistic mesh texture. Surface-aligned Gaussian methods suggest a path toward meshes that better preserve roof planes and boundaries, but satellite deployment should combine them with DSM alignment, georeferenced texture projection, semantic QA, and GIS-ready metadata.

\subsection{Deployment Validation Plan}
Before field deployment, the Agent should be validated on a larger AOI queue that represents the intended operating environment. Each AOI should run the baseline, a conservative semantic setting, and at most one stronger building-focused setting selected by the QA gate. The validation record should store mask coverage, inventory metadata, overlay thumbnails, available DSM errors, and the final policy decision. Acceptance criteria are product-specific: building-focused requests require building-region improvement with a full-scene accuracy bound, while general DSM requests prefer the staged or baseline policy unless both building and full-scene metrics improve.

This validation queue should also separate model performance from product readiness. A scene may have acceptable global MAE but fail a building-focused request, while another scene may justify stronger semantic weighting only after regional DSM validation. Recording these cases gives operators a calibration set for deciding when mask coverage is reliable enough to trigger semantic weighting and gives analysts a repeatable basis for accepting, comparing, or rerunning DSM products across scenes.

For each accepted AOI, the audit card should identify the candidate policies considered, the selected product objective, the mask-coverage and inventory signals used by the gate, and the regional DSM metrics used for verification. Failed or baseline-selected cases should be retained rather than discarded, because they define the operational boundary of the semantic prior. This record is especially important when the same reconstruction pipeline is run across heterogeneous city blocks, where dense roofs, sparse suburbs, vegetation, and shadowed views can produce different policy choices. Over time, these records can support threshold tuning, operator review, and comparison of new reconstruction engines without changing the analyst-facing product contract.

\section{Conclusion}
This paper presents an agentic building-aware foundation-prior workflow for satellite Gaussian Splatting. On mask-vetted scenes, semantic weighting can improve building DSM accuracy; across scenes, the Agentic Reconstruction Controller converts mask QA, inventory metadata, and validation metrics into explicit operating policies. The main application contribution is therefore not a single fixed weight, but a product workflow that reports when semantic priors help and when the baseline is the better reconstruction choice.

The results support a practical principle for urban DSM production: reconstruction systems should expose their policy decisions as clearly as their surfaces. By pairing region-specific DSM metrics with decision records and analyst-facing visual products, the proposed workflow makes neural satellite reconstruction more auditable, easier to triage, and better aligned with GIS deployment needs.


\clearpage
\bibliography{references}

\end{document}